\documentclass[10pt,twocolumn,letterpaper]{article}

\usepackage{iccv}
\usepackage{times}
\usepackage{epsfig}
\usepackage{graphicx}
\usepackage{amsmath}
\usepackage{amssymb}
\usepackage{multirow}
\usepackage{cite}

\usepackage[pagebackref=true,breaklinks=true,letterpaper=true,colorlinks,bookmarks=false]{hyperref}

\iccvfinalcopy 

\def\iccvPaperID{2673} 

\newcommand{\tb}{}

\ificcvfinal\fi

\makeatletter
\newcommand{\thickhline}{%
    \noalign {\ifnum 0=`}\fi \hrule height 1pt
    \futurelet \reserved@a \@xhline
}
\makeatother

\begin{document}

\title{ TransMOT: Spatial-Temporal Graph Transformer for Multiple Object Tracking}

\author{
Peng Chu\textsuperscript{1}\and Jiang Wang\textsuperscript{1} \and Quanzeng You\textsuperscript{1}\and Haibin Ling\textsuperscript{2}\and Zicheng Liu\textsuperscript{1}\\
\small \textsuperscript{1}Microsoft \quad \textsuperscript{2}Stony Brook University\\
{\tt\small \{pengchu, jiangwang, quanzeng.you, zliu\}@microsoft.com, hling@cs.stonybrook.edu}
}

\maketitle
\ificcvfinal\thispagestyle{empty}\fi

\begin{abstract}
    Tracking multiple objects in videos relies on modeling the spatial-temporal interactions of the objects. 
    In this paper, we propose a solution named \emph{TransMOT}, which leverages powerful graph transformers to efficiently model the spatial and temporal interactions among the objects. TransMOT effectively models the interactions of a large number of objects by arranging the trajectories of the tracked objects as a set of sparse weighted graphs, and constructing a spatial graph transformer encoder layer, a temporal transformer encoder layer, and a spatial graph transformer decoder layer based on the graphs. TransMOT is not only more computationally efficient than the traditional Transformer, but it also achieves better tracking accuracy. To further improve the tracking speed and accuracy, we propose a cascade association framework to \tb{handle} low-score detections and long-term \tb{occlusions} that \tb{require} large computational resources to model in TransMOT. The proposed method is evaluated on multiple benchmark datasets including MOT15, MOT16, MOT17, and MOT20, and it achieves state-of-the-art performance on all the datasets.
\end{abstract}

\begin{figure*}[t]
	\centering
	\includegraphics[width=1\linewidth]{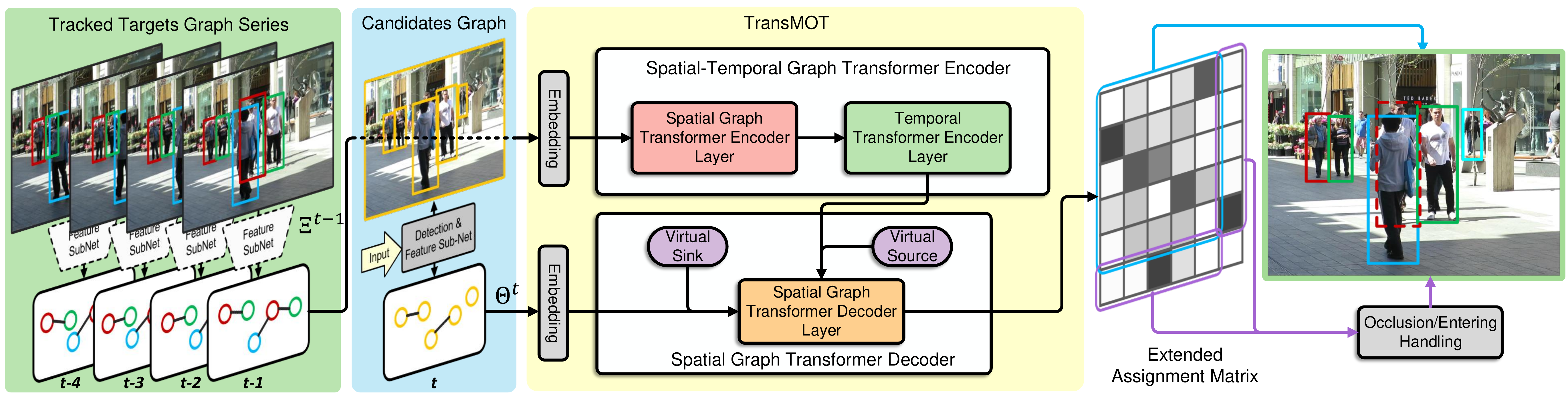}
	\caption{Overview of the proposed TransMOT pipeline for online MOT. The trajectories graph series $\mathbf{\Xi}^{t-1}$ till frame $t-1$ and detection candidates graph $\Theta^t$ at frame $t$ serve as the source and target inputs, respectively, to the spatial-temporal graph transformer.}
	\label{fig:overview}
\end{figure*}
\section{Introduction}

Robust tracking of multiple objects in video is critical for many real-world applications, ranging from vision-based surveillance to autonomous driving vehicles. Most of the recent state-of-the-art Multiple Object Tracking (MOT) \tb{methods} use the tracking-by-detection strategy, where target candidates proposed by an object detector on each frame are associated and connected to form target trajectories~\cite{bae2018confidence,fagot2016improving,keuper2018motion,milan2016multi,pirsiavash2011globally,wang2016tracking,xiang2015learning}.
There are two core tasks in this framework: accurate object detection and robust target association. 
In this paper, we \tb{focus} on building models for robust target association, where successfully modeling the temporal history and appearance of the targets, as well as their spatial-temporal relationships plays an important role.

Traditional models of spatial-temporal relationships usually \tb{rely} on manually \tb{designed} association rules, such as social interaction models or spatial exclusion models~\cite{luo2020multiple}. The recent advances in deep learning \tb{inspire} us to explore to learn the spatial-temporal relationships using deep learning. In particular, the success of Transformer suggests a new paradigm of modeling temporal dependencies through the powerful self-attention mechanism. Recent studies in~\cite{sun2020transtrack,meinhardt2021trackformer} have proved the feasibility of directly modeling spatial-temporal relationships with transformers. However, the tracking performance of transformer-based tracker is not state-of-the-art for several reasons. First, a video contains a large number of objects. Modeling the spatial temporal relationships of these objects with a general Transformer is \tb{ineffective}, 
because it does not take the spatial-temporal structure of the objects into consideration. Second, it requires a lot of computation resources and data to learn a transformer to model long term temporal dependencies. Third, DETR-based object detector used in these works is still not the state-of-the-art for MOT.

In this paper, we propose a novel spatial-temporal graph Transformer for MOT~(TransMOT) to resolve all these issues. In TransMOT, the trajectories of all the tracked targets are arranged as a series of sparse weighted graphs that are constructed using spatial relationships of the targets. Based on these \tb{sparse} graphs, TransMOT builds a spatial graph transformer encoder layer, a temporal transformer encoder layer, and a spatial transformer decoder layer to model the spatial temporal relationships of the objects. It is more computationally efficient during training and inference because of the sparsity of the weighted graph representation.
It is also a more effective model than regular transformer because it exploits the structure of the objects.
We also propose a cascade association framework to handle low-score detections and long-term occlusions. 
By incorporating TransMOT into the cascade association framework, we do not need to learn to associate a large number of low-score detections or model long-term temporal relationships. 
TransMOT can also be combined with different object detectors or visual feature extraction sub-networks to form a unified end-to-end solution so that we can exploit the state-of-the-art object detectors for MOT. Extensive experiments on MOT15, MOT16, MOT17, and MOT20 challenge datasets demonstrate that the proposed approach achieves the best overall performance and establishes new state-of-the-art in comparison with other published works.

In summary, we make following contributions:
\begin{itemize}
\item\vspace{-1.2mm} We propose a spatial-temporal graph Transformer (TransMOT) for effective modeling of the spatial-temporal relationship of the objects for MOT.
\item\vspace{-1.2mm} We design a cascade association framework that can improve TransMOT and other transformer-based trackers by handling low-score detections and long-term occlusion.
\end{itemize}

\section{Related Works}
Most of the recent Multiple Object Tracking (MOT) trackers are based on the tracking-by-detection framework. 
Tracking-by-detection framework generates tracklets by associating object detections in all the frames using matching algorithms such as Hungarian algorithm~\cite{bewley2016simple,fang2018recurrent,huang2008robust}, network flow~\cite{dehghan2015target,zamir2012gmcp,zhang2008global}, and multiple hypotheses tracking~\cite{chen2017enhancing,kim2015multiple}.
Many works solve the association problem by building graphs of object detections across all the frames, such as multi-cuts~\cite{keuper2018motion, tang2016multi} and lifting edges~\cite{tang2017multiple}. However, these methods need to perform computationally expensive global optimization on large graphs, which limits their application {to online} tracking. 

Recently, deep learning-based association algorithm is gaining popularity in MOT~\cite{zhu2018online}. 
In~\cite{ondruska2016deep} and~\cite{milan2017online}, recurrent neural networks (RNN) is explored to solve the association problem using only the motion information. 
In~\cite{chu2019famnet}, a power iteration layer is introduced in the rank-1 tensor approximation framework~\cite{ShiLXH13cvpr} to solve the multi-dimension assignment in MOT. 
\cite{bergmann2019tracking} and~\cite{zhou2020tracking} combine object detection and target association that directly predict target locations in the current frame.
In~\cite{xu2020train}, a differentiable MOT loss is proposed to learn deep Hungarian Net for association.
In~\cite{braso2020learning}, graph convolutional neural network is adopted as a neural solver for MOT, where a dense graph connecting every pair of nodes in different frames is constructed to infer the association.
The proposed TransMOT also constructs a spatial graph for the objects within the same frame, but it exploits the Transformer networking architecture to jointly learn the spatial and temporal relationship of the tracklets and candidates for efficient association.

Transformer has achieved great success in various computer vision tasks, such as detection~\cite{carion2020end} and segmentation~\cite{liang2020polytransform}. In~\cite{yu2020spatio}, Transformer is adopted for trajectory prediction. The studies in~\cite{sun2020transtrack,meinhardt2021trackformer} are the pioneer investigations in applying Transformer in MOT. Both methods use DETR for detection and feature extraction, and model the spatial-temporal relationship of the tracklets and detections using Transformer. The proposed TransMOT framework {utilizes} spatial graph transformer to model spatial relationship of the tracklets and detections, and it factorizes the spatial and temporal transformer encoder for model efficient modeling.

\section{Overview}
We aim at joint detection and tracking multiple objects in videos in an online fashion. Fig.~\ref{fig:overview} illustrates our framework built upon tracking-by-detection framework. The framework maintains a set of $N_{t-1}$ tracklets, each of which represents a tracked object. Each tracklet $\mathbf L^{t-1}_i$ maintains a set of states, such as its past locations $\big\{ \hat{x}_{t'}^{I}\big\}_{t'=t-T}^{t-1}$ and appearance features $\big\{ \hat{f}_{t'}^{I}\big\}_{t'=t-T}^{t-1}$  on the previous $T$ image frames. 
Given a new image frame $I_t$, the online tracking algorithm eliminates the tracklets whose tracked object exits the scene, determines whether any tracked objects are occluded, computes new locations for the existing tracklets $\hat{\mathbf{X}}_{t} = \big\{ \hat{x}^{t}_{i}\big\}^{N_t}_{i=1}$, and generates new tracklets for new objects that {enter} the scene.

As shown in  Fig.~\ref{fig:overview}, our framework contains two major parts: the detection and feature extraction sub-networks, and spatial temporal graph transformer association sub-network. At each frame, the detection and feature extraction sub-networks generate $M_t$ candidate object detection proposals $\mathbf{O}_t = \big\{o^{t}_j\big\}^{M_t}_{j=1}$, as well as visual features for each proposal. The spatial-temporal graph transformer finds the best candidate proposal for each tracklet and models the special events, such as entering, exiting, or occlusion.

For each tracklet $\mathbf{L}^{t-1}_i$, the best matching is obtained through selecting the $o_{t}^j$ maximizing the affinity $\phi(\mathbf {L}^{t-1}_i, o^{t}_j)$, where $\phi(\cdot)$ is a scoring function that computes the affinity of the tracklet state and the candidate. Taking all tracklets into consideration, the problem can be formulated as a constrained optimization problem as 
\vspace{-2mm}\begin{equation}
    \max_{A^t=(a^{t}_{ij})} \ \sum_{i=1}^{N_{t-1}} \sum_{j=1}^{M_t} {a}^{t}_{ij} \phi(\mathbf L^{t-1}_i, o^{t}_j), 
    \label{eq:obj}
\end{equation}
\vspace{-3mm}\begin{equation}
    \text{s.t.} \left\{\!\!
    \begin{array}{rl}
         \sum_i a^{t}_{ij} = 1, & \forall i=1,\dots,N_{t-1} \\
         \sum_j a^{t}_{ij} = 1, & \forall j=1,\dots,M_t\\
         a^{t}_{ij} \in \{0, 1\},&  \forall i=1,\dots,N_{t-1}; j=1,\dots,M_t
    \end{array}
    \right.
\label{eq:st}
\end{equation}
where $A^t=({a}_{ij}^t)$ indicates the association between tracklets $\mathcal L^{t-1}=\{L^{t-1}_i\}_{i=1}^{N_t}$ and detected candidates $\mathcal O^t$. 
Eq.~\ref{eq:st} is used to enforce the assignment constraints.

In order to more effectively model the spatial-temporal relationship between all the tracklets and candidates, the proposed framework rewrites  Eq.~\ref{eq:obj} and Eq.~\ref{eq:st} into a single function ${A}^t =\Phi(\mathcal L^{t-1}, \mathcal O^t)$, where $\mathcal L^{t-1}$ and  $\mathcal O^t$ consist of all the tracklets and candidates, respectively. 

To model the spatial-temporal object correlation, we build a weighted spatial graph $\Theta^t$ for the proposals at the current frame, and a set of weighted spatial graphs $\mathbf{\Xi}^{t-1}=\{\xi^{t- T},\xi^{2},\dots,\xi^{t-1}\}$ of the tracked objects at the previous $T$ frames. 
 The spatial-temporal graph neural network utilizes these graphs to build an efficient spatial-temporal graph transformer that models the relationship between the tracked objects and newly generated proposals. 
 It generates an assignment matrix $\bar{A}^t$ to track the objects and model the special events, such as entering, exiting, or occlusion, as shown in Fig.~\ref{fig:overview}. 
 The assignment matrix is used to update the tracked target while the special events are handled by the post-processing module, which will be explained in Sec.~\ref{sec:cascade}.

The details of the spatial-temporal graph Transformer will be explained in Sec.~\ref{sec:encoder} and Sec.~\ref{sec:decoder}. The two types of training losses to train TransMOT will be elaborated in Sec.~\ref{sec:train}.

\section{TransMOT}

Spatial-temporal graph Transformer for MOT~(TransMOT) uses the graphs $\mathbf{\Xi}^{t-1}$ and $\Theta^t$  to learn a mapping $\Phi(\cdot)$  that models the spatial-temporal correlations, and generates an assignment/mapping matrix $\bar{A}^t$. It contains three parts: a spatial graph transformer encoder layer, a temporal transformer encoder layer, and a spatial graph transformer decoder layer. We propose graph multi-head attention to model spatial relationship of the tracklets and candidates using the self-attention mechanism. It is crucial for both the spatial graph transformer encoder layer and the spatial graph transformer decoder layer.


\subsection{Spatial-Temporal Graph Transformer Encoder}
\label{sec:encoder}

\begin{figure}
	\centering
	\includegraphics[width=0.65\linewidth]{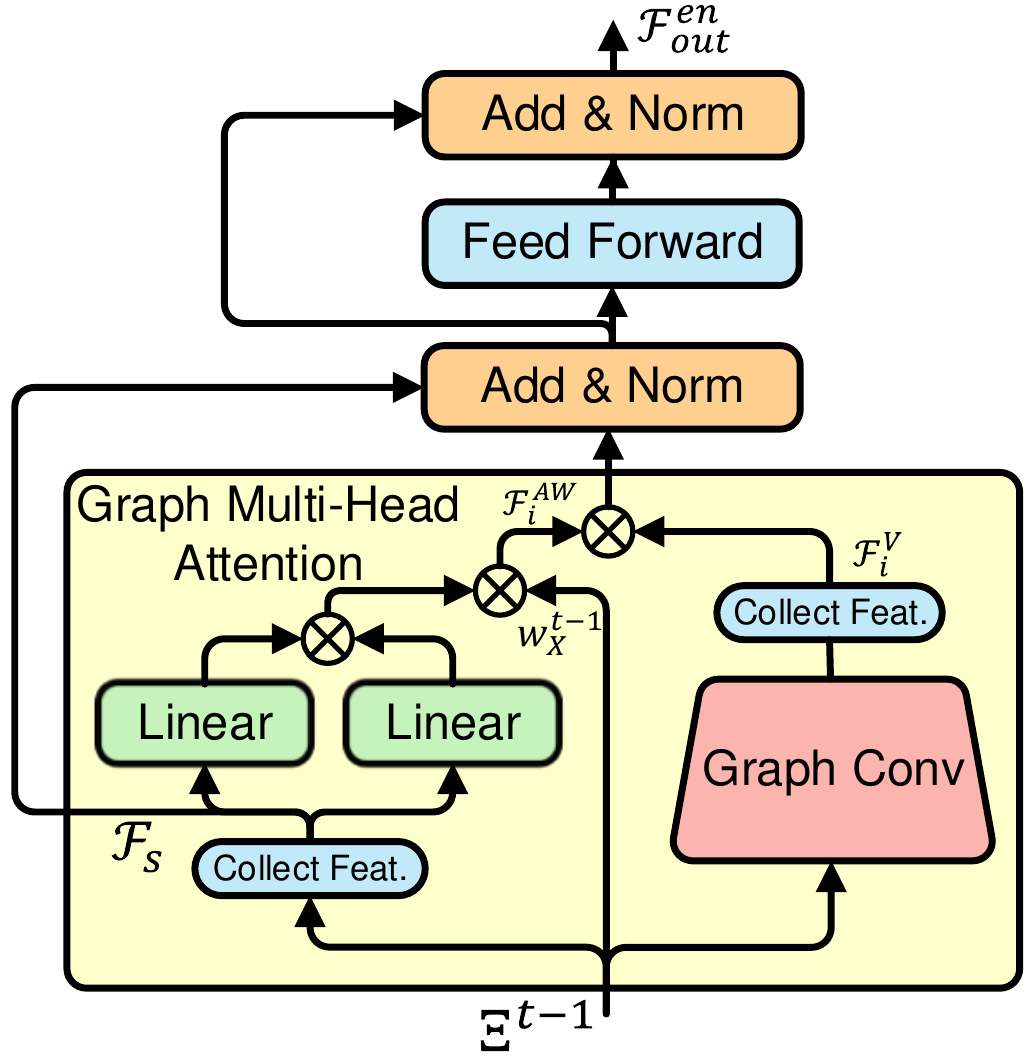}
	\caption{The spatial graph transformer encoder layer. 
	}
	\label{fig:encoder}
\end{figure}

The spatial-temporal graph encoder consists of a spatial-temporal graph transformer encoder layer to model the spatial correlation among tracklets, and a temporal transformer encoder layer to further fuse and encode the spatial and temporal information of the tracklets. We find that by factoring the transformer into spatial and temporal transformers, it makes the model both more accurate and computationally efficient. 

\subsubsection{Spatial Graph Transformer Encoder Layer}
The input of spatial-temporal graph encoder layer is the states of the tracklets for the past $T$ frames.  The tracklet state features are arranged using a sequence of tracklet graphs $\mathbf{\Xi}^{t-1}=\{\xi^{t - T},\xi^{t-T+1},\dots,\xi^{t-1}\}$ , where ${\xi^{t - 1}=G(\{x^{t-1}_i\}, E^{t-1}_X, w^{t-1}_X)}$ is the spatial graph\footnote{We use $G(\cdot)$ to denote a graph.} of the tracklets at frame $t-1$. 
At frame $t-1$, the graph node $x^{t-1}_i$ represents the status of $i$-th tracklet at this frame, two nodes are connected by an edge in $E^{t-1}_X$ if their corresponding bounding boxes have IoU larger than 0, and the edge weight in $w^{t-1}_X$ is set to the IoU. The weight matrix $w^{t-1}_X\in\mathbb{R}^{N_{t-1} \times N_{t-1}}$ is a sparse matrix, whose $(i, j)$ entry is the weight of the edge connecting node $i$ and node $j$, or 0 if they are not connected.

The node features for the tracklets are first embedded through a source embedding layer (a linear layer) independently for each node.  All the node features are arranged into a feature tensor  $\mathcal{F}_{s} \in \mathbb{R}^{N_{t-1}\times T \times D}$, where $D$ is the dimension of the source embedding layer. It is passed into the spatial graph transformer encoder layer together with the graph series as shown in Fig.~\ref{fig:encoder}. Inside the layer, a multi-head graph attention module is utilized to generate self-attention for the input graph series. This module takes feature tensor $\mathcal{F}_s$ and the graph weights $w^{t-1}_X$ to generate self-attention weights for the $i$-th head:
\vspace{-1mm}\begin{equation}
    \mathcal{F}^{AW}_i = \mathrm{softmax}\Big[\varphi(\mathcal{F}_{s}, W_i^Q, W_i^K) \circ w^{t-1}_X\Big],
    \label{eq:att}
    \vspace{-1mm}
\end{equation}
where $\varphi(\cdot)$ is the regular scaled dot-product to obtain attention weights as in~\cite{vaswani2017attention}, and $\circ$ is the element-wise product. It can be understood as computing the spatial graph self-attention for each timestamp independently.

The multi-head graph attention utilizes the graph weights $w^{t-1}_X$ to generate non-zero attention weights only for the tracklets that have spatial interactions, because the tracklets that are far way from each other usually have very little interaction in practice. By focusing its attention on a much smaller subset, the spatial graph transformer encoder layer models the interactions more effectively, and runs faster during training and inference.

We also apply graph convolution instead of the linear layer to aggregate information from neighboring nodes. After the graph convolution layer, the node features are collected to form a value tensor $\mathcal{F}^{V}_i$. Combined with the attention weights in Eq.~\ref{eq:att}, the graph multi-head attention weighted feature tensor can be written as 
\begin{equation}
    \mathcal{F}_{att}^{en} = \mathrm{Concate}(\{\mathcal{F}^{AW}_i\otimes \mathcal{F}^{V}_i\})\otimes W^O,
    \nonumber
\end{equation}
where $\{\cdot\}$ iterates and aggregates the outputs from all the attention heads, $\otimes$ is the tensor mode product\footnote{It performs matrix product of each slice of right and left tensors along the dimension sharing the same length.}

The attention weighted feature tensor is projected through a linear feed forward and a normalization layer to get the final output of the spatial graph transformer encoder layer. 

\subsubsection{Temporal Transformer Encoder Layer}

The features of the tracklets are further encoded by a temporal transformer encoder layer. The temporal transformer encoder layer transposes the first two dimension of the output tensor from the spatial graph transformer encoder, resulting in a tensor $\mathcal{F}_{tmp}^{en} \in \mathbb{R}^{T \times N_{t-1} \times D}$. 
The temporal transformer encoder layer employs a standard Transformer encoder layer over the temporal dimension for each tracklets independently. 
It calculates the self-attention weights along the temporal dimension, and computes the temporal attention-weighted feature tensor for the tracklets.

The output of the temporal transformer encoder layer is the final output of the spatial-temporal graph transformer encoder $\mathcal{F}_{out}^{en}$.  

\subsection{Spatial Graph Transformer Decoder}
\label{sec:decoder}

\begin{figure}
	\centering
	\includegraphics[width=0.8\linewidth]{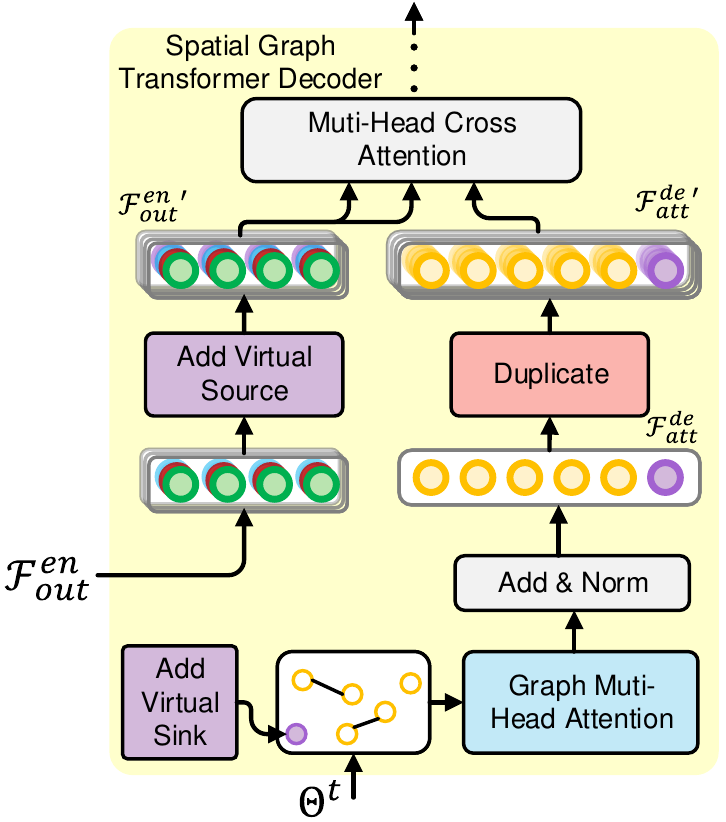}
	\caption{Illustration of the spatial graph transformer decoder.}
	\label{fig:decoder}
\end{figure}

The spatial graph transformer decoder produces extended assignment matrix $\bar{A}^t$ from the candidate graph ${\Theta^{t}=G(\{o^{t}_j\}, E^{t}_O, w^{t}_O})$ and the output of the spatial-temporal transformer encoder $\mathcal{F}_{out}^{en}$.  
The candidate graph is constructed similarly to the tracklet graphs in Sec.~\ref{sec:encoder}. Each node $o^{t}_j$ represents a candidate in frame $t$. \tb{Two nodes} are connected only if their bounding box's IoU is larger than zero, and the weight of the edge is set to the IoU. 
Besides the nodes representing the real candidates, a virtual sink node is added to the graph. The virtual sink node is responsible for exiting or occlusion events of any tracklet in the current frame. In particular, a node with a set of learnable embedding $f_{snk} \in \mathbb{R}^D$ is added to the $\Theta^t$. The virtual sink node is connected to all the other nodes with weight $0.5$.

Similar as the encoder in Sec.~\ref{sec:encoder}, the candidate node features of input graph are embedded and collected. The $f_{snk}$ is appended to the embedded feature set such that $\mathcal{F}_{tgt}^{de} \in \mathbb{R}^{(M_t + 1) \times 1\times D}$. 
The spatial graph decoder first uses graph multi-head attention to encode the node features that is similar to the one Sec.~\ref{sec:encoder}, shown in Fig.~\ref{fig:decoder}. 
We denote the attention weighted candidate node features as $\mathcal{F}_{att}^{de} \in \mathbb{R}^{(M_t + 1) \times 1\times D}$.

For the tracklet embedding $\mathcal{F}^{en}_{out}$ generated by the spatial-temporal graph transformer encoder, we add a virtual source to handle the candidates that initiate a new tracklet in the current frame $t$ to form an extended tracklet embedding $\mathcal{F}_{out}^{en'} \in \mathbb{R}^{T \times (N_{t-1} + 1) \times D}$. The embedding of the virtual source is a learnable parameter.
Note that we only add one virtual source node compared to multiple virtual source nodes in Transform-based MOT trackers, because we find adding one virtual source node yields comparable performance as adding multiple virtual source nodes while achieving better computational efficiency. $\mathcal{F}_{att}^{de}$ is duplicated $N_{t-1} + 1$ times such that $\mathcal{F}_{att}^{de} \to \mathcal{F}_{att}^{de'} \in \mathbb{R}^{(M_{t} + 1) \times (N_{t-1} + 1) \times D}$. Multi-head cross attention is calculated for $\mathcal{F}_{att}^{de'}$ and $\mathcal{F}^{en'}_{out}$ to generate 
unnormalized attention weights. 
The output is passed through a feed forward layer and a normalization layer to generate the output tensor 
$\mathbb{R}^{(M_{t} + 1) \times (N_{t-1} + 1) \times D}$ that corresponds to the matching between the tracklets and the candidates.

The output of the spatial graph decoder can be passed through a linear layer and a Softmax layer to generate the assignment matrix $\bar{A}^t \in \mathbb{R}^{(M_t + 1)\times (N_{t-1} + 1)}$.

\subsection{Training}
\label{sec:train}

The TransMOT is trained end-to-end with the guidance of the groundtruth extended assignment matrix. The constraints in Eq.~\ref{eq:st} need to be relaxed to allow efficient optimization. We relax the constraints so that a detection candidate is always associated with a tracklet or a virtual source, while a tracklet can be associated with multiple candidates. In this way, Eq.~\ref{eq:st} can be relaxed as:
\vspace{-3mm}\begin{equation}
    \vspace{-3mm}
    \text{s.t.} \quad \sum_j^{N_{t-1} + 1} \bar{a}^{t}_{ij} = 1, i \in [1, M_t], \bar{a}^{t}_{ij} \in \{0, 1\}.
\nonumber
\label{eq:st_reduced}
\end{equation}
As a result, a row of the assignment matrix can be treated as a probability distribution over a total of $N_{t-1} + 1$ categories, and we use the cross-entropy loss to optimize the network. 

In each training iteration, a continuous sequence of $T+1$ frames are randomly sampled from the training set. The bounding boxes and their corresponding IDs are collected from each frame. The groundtruth bounding boxes are then replaced by the bounding boxes generated from the object detector by matching their IoUs. In this way, the TransMOT will be more robust to detection noise. For all bounding boxes in a frame, their IDs are remapped to $\{0, 1, \cdots, N_{t-1}\}$ indicating whether they are matched to a tracklet or a virtual source. 

For the rows that correspond to actual tracklets, a cross-entropy loss is utilized, as mentioned above.
The last row of $\bar{A}^t$ represents the virtual sink, and it may be matched to multiple tracklets. 
Thus, a multi-label soft margin loss is employed to optimize this part separately. 

In summary, the overall training loss can be written as
\vspace{-2mm}\begin{equation}
\vspace{-2mm}
\begin{aligned}
    \mathcal{L} = &-\frac{1}{M_t}\sum_{m=1}^{M_t}y_m \mathrm{log}({\bar{\mathbf{a}}_m}) \\
    &+ \frac{\lambda}{N_{t-1}}\sum_{n=1}^{N_{t-1}}y_n^{snk}\mathrm{log}\Big(\frac{1}{1+e^{-a_{n}'}}\Big)\\
    &+ \frac{\lambda}{N_{t-1}}\sum_{n=1}^{N_{t-1}}(1-y_n^{snk})\mathrm{log}\Big(\frac{e^{-a_n'}}{1 +e^{-a_n'}}\Big),
    \nonumber
\end{aligned}
\end{equation}
where $y_m$ and $y_n^{snk}$ are IDs of the detection candidates and the virtual sink respectively, $\bar{\mathbf{a}}_m$ is the row element of $\bar{A}^t$, $\bar{\mathbf{a}}_{M_t + 1} = \{a_n'\}$, and $\lambda$ is a weighting coefficient.

\subsection{Cascade Association Framework}
\label{sec:cascade}

\begin{figure}
	\centering
	\includegraphics[width=1\linewidth]{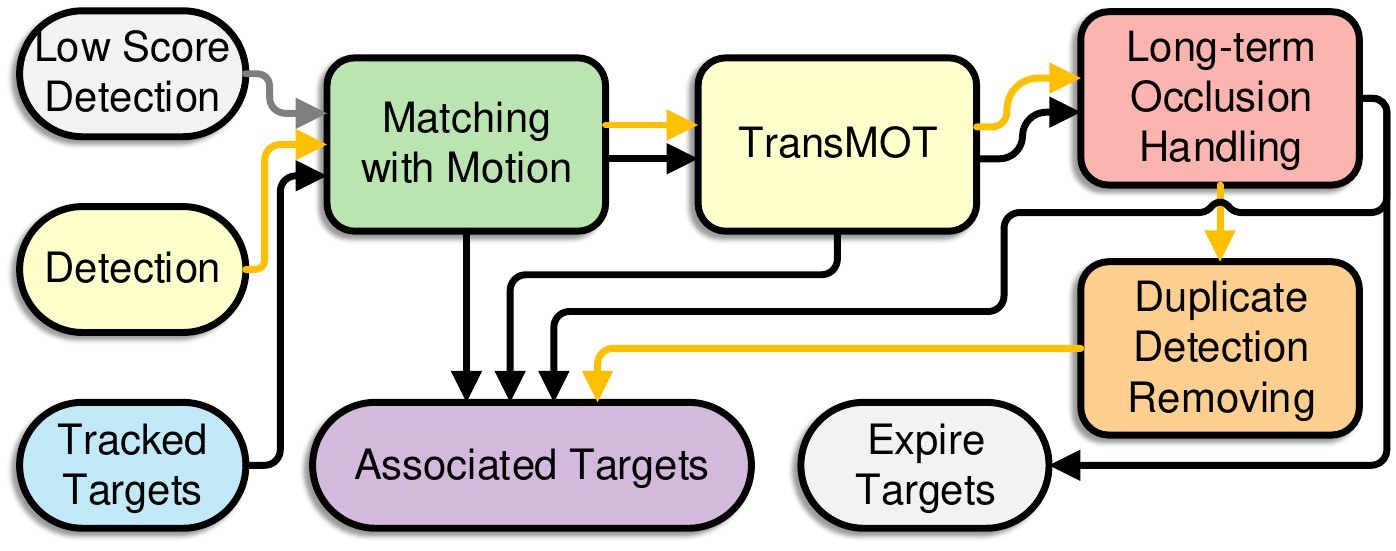}
	\caption{Illustration of the cascade association framework based tracking system.}
	\label{fig:cascade}
\end{figure}

Although the spatial-temporal graph transformer network can effectively model the spatial-temporal relationship between tracklets and \tb{object} candidates, we can achieve better inference speed and tracking accuracy by incorporating it in a three-stage cascade association framework. The illustration of the cascade association framework is shown in Fig.~\ref{fig:cascade}.

The first stage matches and filters the low confidence candidate boxes using motion information. 
In particular, a Kalman Filter predicts the bounding boxes of the robustly tracked tracklets in the current frames.
These tracklets must be successfully and continuously associated on the past $K_r$ frames. 
The IoU between the predicted bounding boxes and the candidate boxes are utilized as the association score.
We match the predicted boxes and the candidate boxes with Hungarian algorithm.
Only the matched pairs with IoU larger than $\tau_M$ are associated.
The rest of the candidate boxes whose confidence scores are lower than a threshold is filtered.
This stage is mainly a speed optimization, because it removes the candidate boxes and tracklets that can be easily matched or filtered, and leaves the more challenging tracklets and candidate boxes for TransMOT for further association.

In the second stage, TransMOT calculates the extended assignment matrix for the rest of the tracklets and candidates. 
The upper left part of the extended assignment matrix $\bar{A}^t$ denotes as $\hat{A} \in \mathbb{R}^{M_t\times N_{t-1}}$ determines the matching of actual tracklets and candidate boxes. Since the elements of $\hat{A}\in [0, 1]$ is a soft assignment, we apply bipartite matching algorithm to 
generate the actual matching. Similarly, only the pairs with assignment scores larger than a threshold will be matched at this stage. After the matching, the unresolved candidate boxes will be matched again to recover occlusion or initiate new targets in the third stage. 

The third stage is the Long-Term Occlusion and Duplicated detection handling module in Fig.~\ref{fig:cascade}. 
Since TransMOT only models the tracklets of the previous $T$ frames, the tracklets that are occluded in the previous $T$ frames are not matched.
For these tracklets, we store their visual features and the bounding box coordinates at the latest frame when they were visible, and use them to calculate the association cost of a tracklet and a candidate detection.
The association cost is defined as the addition of the Euclidean distance of the visual features and the normalized top distance of the occluded tracklets and candidate boxes.
\begin{equation}
    \mathcal{D}_{top} = \left\Vert\left(u_i + \frac{w_i}{2} - u_j -\frac{w_j}{2}, v_i - v_j\rm\right) \right\Vert \Bigm/ h_i,
    \nonumber
\end{equation}
where $[u,v]$ indicates the left upper corner of the bounding box and $[w,h]$ indicates its size\footnote{The notation $w$ should not be confused with the weight $w$ in Sec.~\ref{sec:encoder}.}. 

Duplicate detection handling removes the unmatched detection candidates that might be duplicate of the detections that are already matched to a tracklet.
In this step, the un-associated candidates are matched to all the associated candidates.
The association cost for this step is the bounding box intersection between the associated box and un-associated box over the area of the un-associated candidates. 
This cost is chosen so that a candidate box that is a sub-box of a matched candidate is removed, because it is very likely that this box is a duplicate detection.
All the matched candidate detection at this step will be removed directly. 

Finally, each of the remaining candidates is initialized as a new tracklet, and the unresolved tracklets that have not been updated for more than $K_p$ frames \tb{are} removed. 
The tracklets that are not updated for less than $K_p$ frames are set to ``occluded'' state.

\begin{table}[!t]
	\footnotesize
	\begin{center}

		\begin{tabular}{c@{\hskip 0.1mm}|@{\hskip 1.5mm}c@{\hskip 1.5mm}c@{\hskip 1.5mm}c@{\hskip 1.5mm}c@{\hskip 1.5mm}c@{\hskip 1.5mm}c@{\hskip 1.5mm}c@{\hskip 1.5mm}c@{\hskip 1.5mm}}
			\hline\thickhline
			 Method & IDF1 & MOTA & MT & ML$\downarrow$ & FP$\downarrow$ & FN$\downarrow$ & IDS$\downarrow$\\
			\hline
			DMT\cite{kim2016cdt} & 49.2 & 44.5 & 34.7\% &22.1\% & 8,088 & 25,335 & 684 \\
			
			
			
			TubeTK \cite{Pang_2020_CVPR}& 53.1 & 58.4 & 39.3\%& 18.0\%& \textbf{5,756}& 18,961 &854\\
			 
			CDADDAL\cite{bae2018confidence} & 54.1 & 51.3 & 36.3\% &22.2\% & 7,110 & 22,271 & 544 \\
			
 			TRID \cite{manen2017pathtrack}& 61.0& 55.7 & 40.6\%& 25.8\%& 6,273 & 20,611 & 351 \\
			 
			RAR15 \cite{fang2018recurrent}& 61.3 & 56.5 & 45.1\%& 14.6\%& 9,386& 16,921 &428\\
			 
			GSDT \cite{wang2020joint}& 64.6 & \textbf{60.7} & 47.0\%& \textbf{10.5\%} & 7,334	 & 16,358 & 477\\
			Fair \cite{zhang2020fairmot}& 64.7 & 60.6 & 47.6\%& 11.0\% & 7,854	 & 15,785 & 591\\
			\hline

			 \textbf{TransMOT} &\textbf{66.0}&57.0&\textbf{64.5\%} &17.8\% &12,454  &\textbf{13,725} & \textbf{244}\\
			\hline\thickhline
		\end{tabular}

	\end{center}
	\vspace{-2mm}
			\caption{Tracking Performance on the MOT15 benchmark test set private detection track. Best in bold.}
		\label{table:res15p}
\end{table}

\begin{table}
	\footnotesize
	\begin{center}
		
		\begin{tabular}{c@{\hskip 0.1mm}|@{\hskip 1.5mm}c@{\hskip 1.5mm}c@{\hskip 1.5mm}c@{\hskip 1.5mm}c@{\hskip 1.5mm}c@{\hskip 1.5mm}c@{\hskip 1.5mm}c@{\hskip 1.5mm}c@{\hskip 1.5mm}}
			\hline\thickhline
			 Method & IDF1 & MOTA & MT & ML$\downarrow$ & FP$\downarrow$ & FN$\downarrow$ & IDS$\downarrow$\\
			\hline
			
			IoU\cite{bochinski2017high} & 46.9 & 57.1 & 23.6\% &32.9\% &5,702& 70,278 &2,167 \\
			
			CTracker\cite{peng2020chained} & 57.2 & 67.6 & 32.9\% &23.1\% &8,934 &48,305 &1,897 \\
			
			
			
			LMCNN \cite{babaee2019dual}& 61.2& 67.4 & 38.2\%& 19.2\%& 10,109 &48,435 &931\\
			
			DeepSort \cite{wojke2017simple} & 62.2 & 61.4 & 32.8\%& 18.2\% & 12,852 & 56,668 & 781\\
			
			FUFET \cite{shan2020fgagt}& 68.6 & 76.5 & 52.8\%& \textbf{12.3}\%& 12,878& 28,982 &1,026\\
			
			LMP \cite{tang2017multiple}& 70.1 & 71.0 & 46.9\%& 21.9\%& \textbf{7,880}& 44,564 &\textbf{434}\\
			CSTrack \cite{liang2020rethinking}& 73.3 & 75.6 & 42.8\%& 16.5\%& 9,646	 & 33,777 & 1,121\\
			\hline

			 \textbf{TransMOT} &\textbf{76.8}&\textbf{76.7}&\textbf{56.5\%} &19.2\% &14,999  &\textbf{26,967} & 517\\
			\hline\thickhline
		\end{tabular}
	\end{center}
	\vspace{-2mm}
	\caption{Tracking Performance on the MOT16 benchmark test set private detection track. Best in bold.}
	\label{table:res16p}
\end{table}

\section{Experiments}

We conduct extensive experiments on four standard MOT challenge datasets for pedestrian tracking: MOT15\cite{MOTChallenge2015}, MOT16, MOT17, and MOT20~\cite{MOT16}. The proposed TransMOT based tracking framework is evaluated on both public and private detection tracks.

\subsection{Experiment Setting and Implementation Details}

The proposed approach is implemented in PyTorch, and the training and inference are performed on a machine with a 10 cores CPU@3.60GHz and an Nvidia Tesla V100 GPU. 
We set the number of frames for tracklets $T=5$, the feature embedding dimension $D=1024$, and the number of heads for all the multi-headed attention in spatial and temporal transformers to 8. 
For graph multi-head attention module, a single layer of ChebConv from~\cite{defferrard2016convolutional} with neighboring distance of 2 is adopted. 
The node features for an object at a frame are the concatenation of its visual features and normalized bounding box coordinates.
During training, we use vanilla SGD with an initial learning rate of 0.0015. 
For all the experiments in Sec.~\ref{sec:res}, we use the training dataset from~\cite{lin2020human} to train our TransMOT model.
During inference, $\tau_M$ is set to 0.75 for selecting confident associations. 
$K_r$ and $K_p$ are set to 15 and 50 respectively.

We trained a YOLOv5~\cite{yolov5} detector model with 407 layers for 300 epochs on the combination of CrowdHuman dataset~\cite{shao2018crowdhuman} and the training sets of MOT17/MOT20.
The SiamFC network~\cite{sadeghian2017tracking} pretrained on the ILSVRC15 dataset is adopted as our visual feature extraction sub-network. 
The maximum input image dimension of the tracking pipeline is set to 1920. The detector runs at 15.4 \textit{fps} on our machine, while the TransMOT and visual feature extraction sub-network run at 24.5 \textit{fps}. The whole tracking pipeline runs at 9.6 \textit{fps}. We also experimented with using TransTrack~\cite{sun2020transtrack} as our detection and feature extraction sub-network, as well as other visual features. These comparisons will be compared in the MOT16/17 and ablation parts of Sec.~\ref{sec:res}. 

To evaluate the performance of the proposed method, the standard ID score metrics~\cite{ristani2016performance} and CLEAR MOT metrics \cite{bernardin2008evaluating} are reported.
ID score metrics calculate the trajectory level ID precision (IDP), ID recall (IDR), and the IDF1 scores.
CLEAR MOT metrics include multiple object tracking precision (MOTP) and multiple object tracking accuracy (MOTA) that combine false positives (FP), false negatives (FN) and the identity switches (IDS). 
The percentage of mostly tracked targets (MT) and the percentage of mostly lost targets (ML) are also reported.



\begin{table}
	\footnotesize
	\begin{center}

		\begin{tabular}{@{\hskip 0.2mm}c@{\hskip 0.5mm}|@{\hskip 0.1mm}c@{\hskip 0.1mm}c@{\hskip 1.5mm}c@{\hskip 0.1mm}c@{\hskip 1.5mm}c@{\hskip 1.5mm}c@{\hskip 1.5mm}c@{\hskip 1.5mm}c@{\hskip 1.5mm}}
			\hline\thickhline
			 & Method & IDF1 & MOTA & MT & ML$\downarrow$ & FP$\downarrow$ & FN$\downarrow$ & IDS$\downarrow$\\
			\hline
			\multirow{7}{*}{\rotatebox{90}{Public Detection}}
			
			&TrctrD\cite{xu2020train} & 53.8 & 53.7 & 19.4\% & 36.6\% & 11,731 & 247,447 & 1,947 \\
			
			&Tracktor\cite{bergmann2019tracking} & 55.1 & 56.3 & 21.1\% & 35.3\% & 8,866 & 235,449 & 1,987 \\

			&CTTrack \cite{zhou2020tracking}& 59.6 & 61.5 & 26.4\%& 31.9\%& 14,076 & 200,672 & 2,583\\
			
			&TrackFormer\cite{meinhardt2021trackformer} & 59.8 & 61.8 & 35.4\% & \textbf{21.1\%} & 35,226 & 177,270 & 2,982 \\
			
			&MPNTrack\cite{braso2020learning} & 61.7 & 58.8 & 28.8\% & 33.5\% & 17,413 & 213,594 & 1,185 \\
			
			&LifT \cite{hornakova2020lifted}& 65.6& 60.5& 27.0\%& 33.6\%& 14,966 & 206,619 & 1,189\\
			
			&MAT \cite{han2020mat}& 69.2& 67.1& \textbf{38.9\%} & 26.4\%& 22,756 & \textbf{161,547} & 1,279\\
			
			\cline{2-9}
			&\textbf{TransMOT-P} &\textbf{72.2}&\textbf{68.7}&33.5\% &31.0\% & \textbf{8,078} & 167,602 & \textbf{1,014}\\
			\hline\hline
			\multirow{7}{*}{\rotatebox{90}{Private Detection}}
			&DAN\cite{sun2019deep} & 49.5 & 52.4 & 21.4\% &30.7\% &25,423 &234,592 &8,431 \\
			
			&TransTrack\cite{sun2020transtrack} & 56.9 & 65.8 & 32.2\% &21.8\% &24,000 &163,683 &5,355 \\
			
			&TubeTK \cite{Pang_2020_CVPR}& 58.6& 63.0& 31.2\%& 19.9\%& 27,060 &177,483 &5,727\\
			
			&CTTrack \cite{zhou2020tracking}& 64.7 & 67.8& 34.6\%& 24.6\%& \textbf{18,498}& 160,332 &6,102\\
			
			&Fair \cite{zhang2020fairmot}& 72.3& 73.7& 43.2\%& 17.3\%& 27,507 &117,477& 8,073\\
			
			
			\cline{2-9}
			&\textbf{TransMOT-D} & 66.9 & 68.8 &35.8\% & 31.7\% &	26,670 &147,690 & \textbf{1,797}\\
			& \textbf{TransMOT} &\textbf{75.1}&\textbf{76.7}&\textbf{51.0\%} & \textbf{16.4\%} &36,231 &\textbf{93,150} &2,346\\
			\hline\thickhline
		\end{tabular}

	\end{center}
		\vspace{-2mm}
	\caption{Tracking Performance on the MOT17 benchmark test set. Best in bold.}
    \label{table:res17}

\end{table}

\begin{table}
	\footnotesize
	\begin{center}
	
			\begin{tabular}{@{\hskip 0.2mm}c@{\hskip 0.5mm}|@{\hskip 0.1mm}c@{\hskip 0.1mm}c@{\hskip 1.5mm}c@{\hskip 0.1mm}c@{\hskip 1.5mm}c@{\hskip 1.5mm}c@{\hskip 1.5mm}c@{\hskip 1.5mm}c@{\hskip 1.5mm}}
			\hline\thickhline
			 & Method & IDF1 & MOTA & MT & ML$\downarrow$ & FP$\downarrow$ & FN$\downarrow$ & IDS$\downarrow$\\
			\hline
			\multirow{5}{*}{\rotatebox{90}{Public Det.}}
			&SORT*\cite{bewley2016simple} & 45.1 & 42.7 & 16.7\% & 26.2\% & 27,521 & 264,694	& 4,470 \\
			&Tracktor\cite{bergmann2019tracking} & 52.7 & 52.6 & 29.4\% & 26.7\% & 6,930 & 236,680 & 1,648 \\

			&MPNTrack\cite{braso2020learning} & 59.1 & 57.6 & 38.2\% & 22.5\% & 16,953 & 201,384 & 1,210 \\
			
			&LPCMOT \cite{dai2021learning}& 62.5& 56.3& 34.1\%& 25.2\%& \textbf{11,726} & 213,056 & 1,562 \\

			\cline{2-9}
			&\textbf{TransMOT-P} &\textbf{74.3}&\textbf{73.1}& \textbf{54.3\%} & \textbf{14.6\%} & 12,366 & \textbf{125,665} & \textbf{1,042}\\
			\hline\hline
			\multirow{5}{*}{\rotatebox{90}{Private Det.}}
			&MLT\cite{zhang2020multiplex} & 54.6 & 48.9 & 30.9\% &22.1\% & 45,660 & 216,803 & 2,187 \\
			
			&GSDT \cite{wang2020joint}& 67.5 & 67.1 & 53.1\%& 13.2\%& 31,913 & 135,409 & 3,131\\
			
			 &Fair \cite{zhang2020fairmot}& 67.3 & 61.8 & 68.8\%& \textbf{7.6}\%& 103,440 & 88,901 & 5,243\\
			
			 
			 &CSTrack \cite{liang2020rethinking}& 68.6 & 66.6 & 50.4\%& 15.5\%& \textbf{25,404} & 144,358 & 3,196\\
			
			\cline{2-9}
			
			 &\textbf{TransMOT} &\textbf{75.2}&\textbf{77.5}&\textbf{70.7\%} &9.1\% &34,201  &\textbf{80,788} & \textbf{1,615} \\
			\hline\thickhline
		\end{tabular}

			
			
			
			 
			
			

	\end{center}
	\vspace{-2mm}
	\caption{Tracking Performance on the MOT20 benchmark test set. Best in bold. Method marked with * in public detection track does not use public detection filtering mechanism. It might achieve better tracking accuracy if the mechanism is employed.}
	\label{table:res20p}
\end{table}
\vspace{-2mm}

\subsection{Evaluation Results}
\label{sec:res}
\begin{figure*}
	\small
	
	\centering
	\begin{tabular}
		{@{\hspace{0.3mm}}c@{\hspace{0.5mm}}c@{\hspace{.5mm}}c@{\hspace{.55mm}}c@{\hspace{.5mm}}}%
		MOT15: Venice-1 & MOT15: AVG-TownCentre & MOT16-03 & MOT16-07
		\\
		\includegraphics[width=0.246\linewidth]{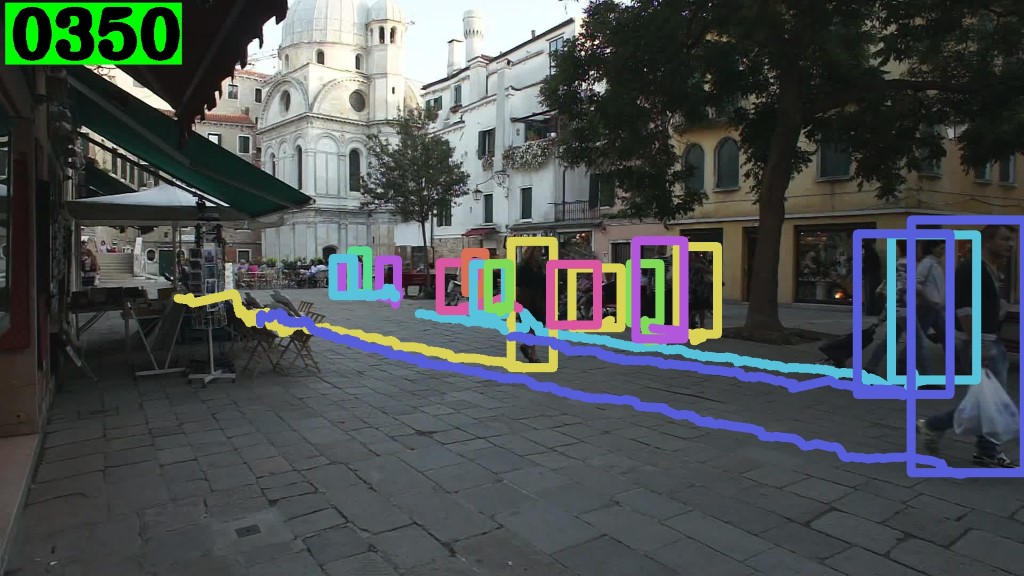}    & 
		\includegraphics[width=0.246\linewidth]{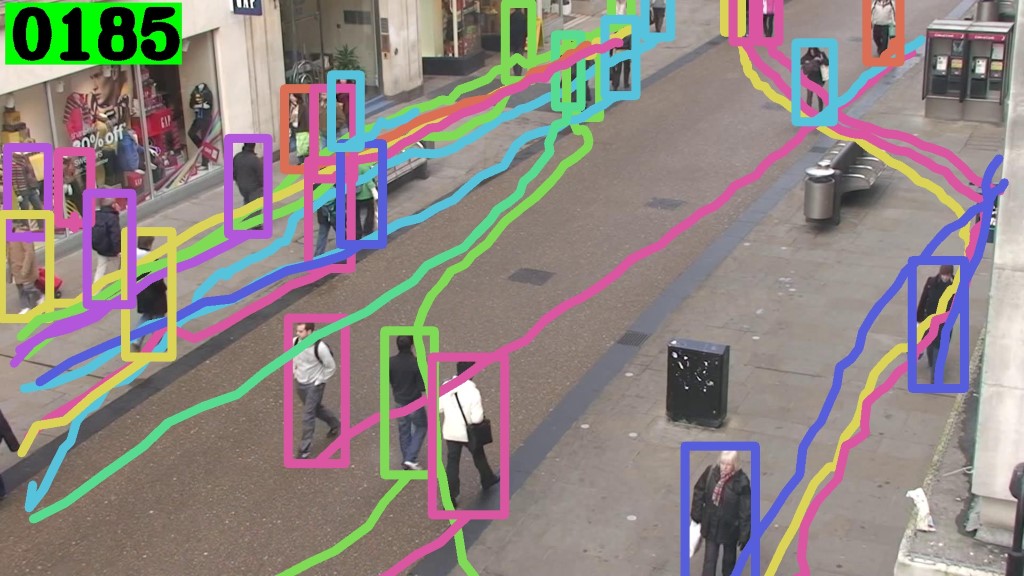}    & 
		\includegraphics[width=0.246\linewidth]{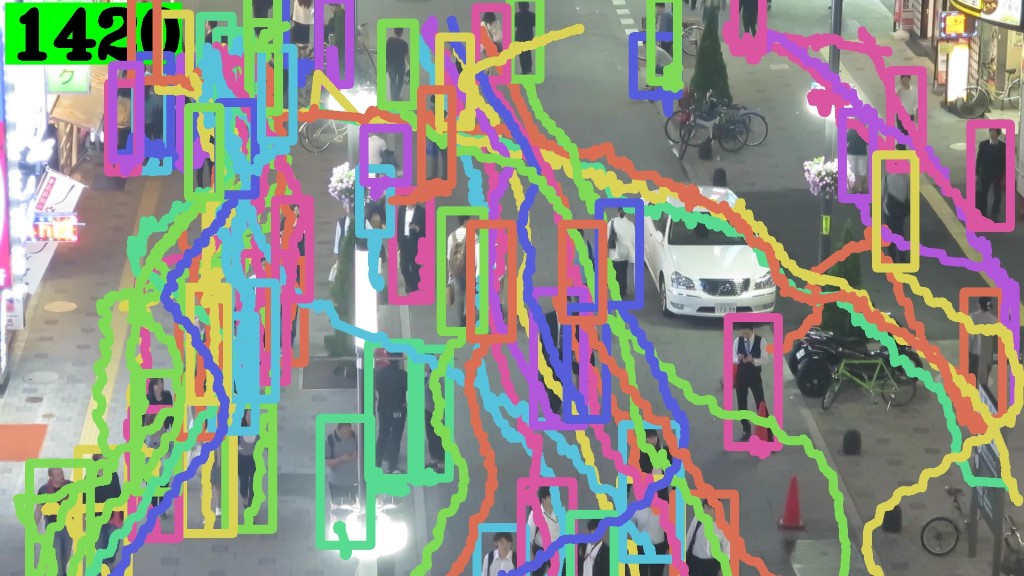}    &
		\includegraphics[width=0.246\linewidth]{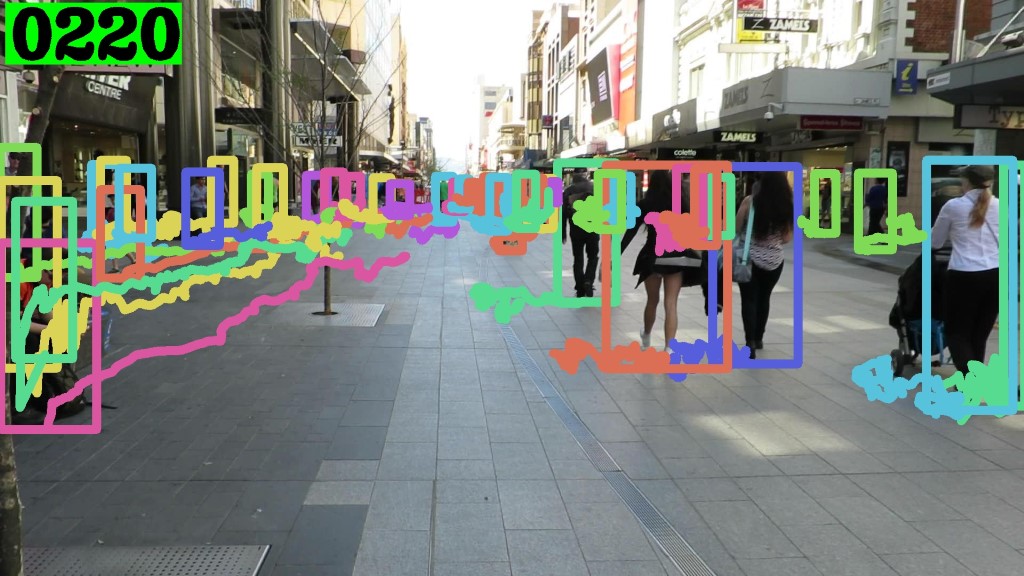}
		\\
		 \end{tabular}

		 \vspace{-1mm}
		 \begin{tabular}
		{@{\hspace{0.3mm}}c@{\hspace{0.4mm}}c@{\hspace{.4mm}}c@{\hspace{.6mm}}c@{\hspace{.4mm}}}%
		\includegraphics[width=0.23\linewidth]{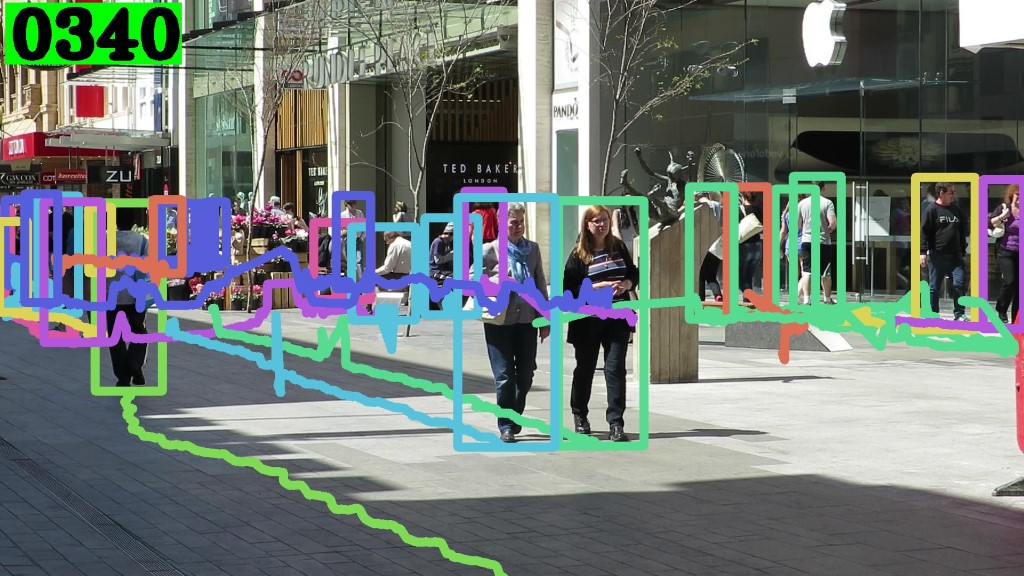}   & 
		\includegraphics[width=0.23\linewidth]{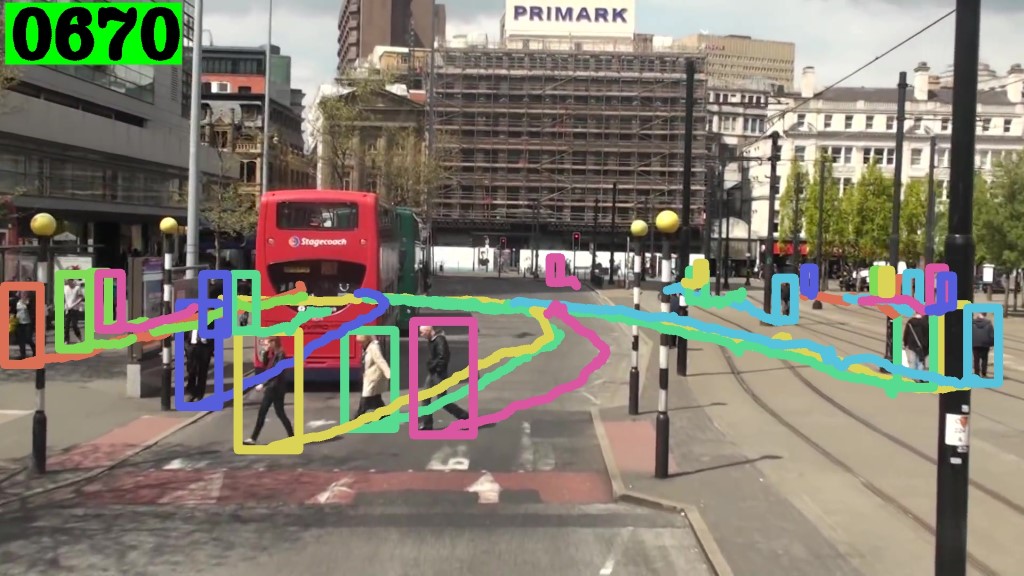}   &
		\includegraphics[width=0.186\linewidth]{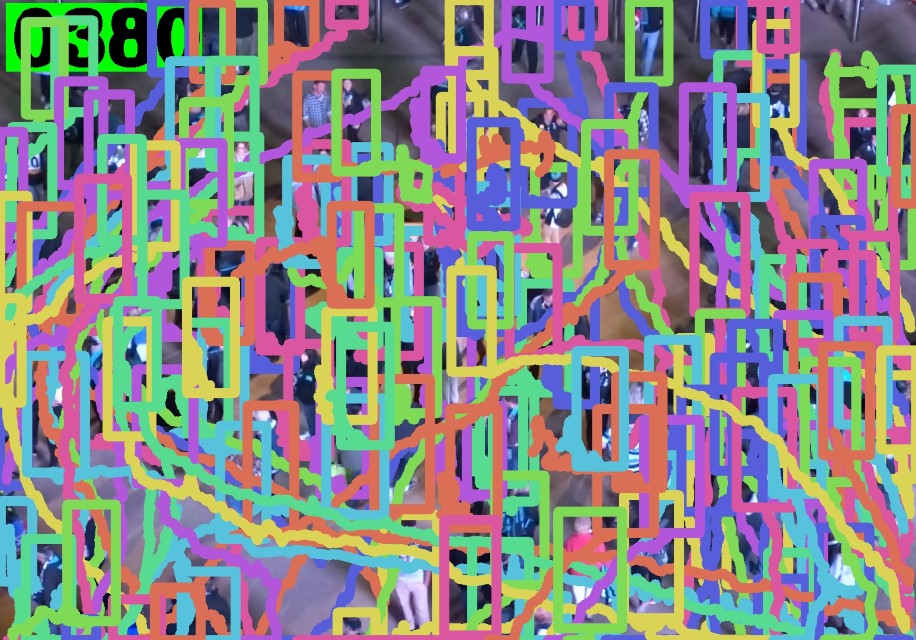}   &
		\includegraphics[width=0.34\linewidth]{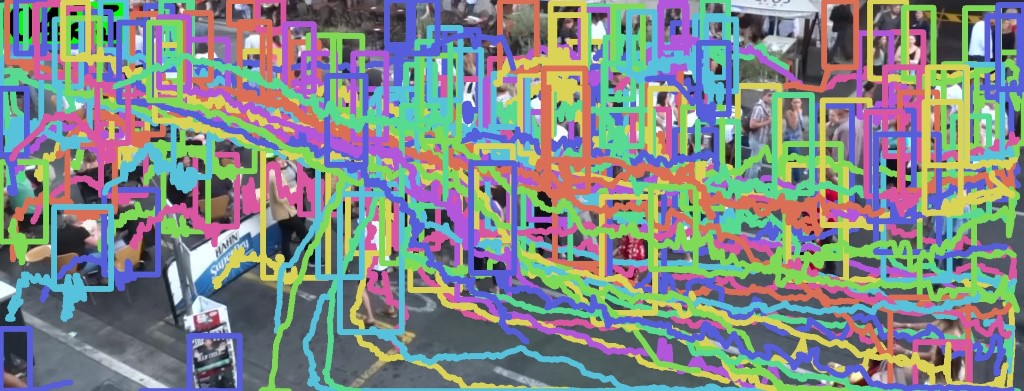}\\
		MOT17-08 & MOT17-14 & MOT20-04 & MOT20-06\\
		\end{tabular}
	
	\caption{ Results visualization of selected sequences in MOT15, MOT16, MOT17, and MOT20.}
	\label{fig:vis}
\end{figure*}

\noindent\textbf{MOT15.}
MOT15~\cite{MOTChallenge2015} contains 22 different indoor and outdoor scenes for pedestrian tracking. 
The 22 sequences are collected from several public and private datasets, and they were recorded with different camera motion, camera angles and imaging conditions. 
The dataset is equally split for training and testing. 
We report the quantitative results of the proposed method on the private detection track in Tab.~\ref{table:res15p}, and the visualizations of the tracking results on selected videos are shown in Fig.~\ref{fig:vis}. 
TransMOT achieves state-of-the-art performance in metrics IDF1, MT, FN, and IDS. 
The relatively lower MOTA score on this dataset is caused by the high FP rate, because not all the objects are exhaustively annotated for some testing sequences.

\noindent\textbf{MOT16/17.}
MOT16 and MOT17~\cite{MOT16} contain the same 14 videos for pedestrians tracking.
MOT17 has more accurate ground truth annotations compared to MOT16 dataset. 
MOT17 also evaluates the effect of object detection quality on trackers, by providing three pretrained object detectors using  DPM~\cite{felzenszwalb2010object}, Faster-RCNN~\cite{ren2015faster} and SDP~\cite{yang2016exploit}. 
We report the performance and comparisons with the state-of-the-art methods on the private detection track of MOT16 in Tab.~\ref{table:res16p}. 
Our approach outperforms all other published trackers using the private detector in both IDF1 and MOTA metrics. 

In MOT17, for a more complete comparison, we configure TransMOT as two additional settings: TransMOT-P and TransMOT-D. 
TransMOT-P uses the public detection results, and it follows the filtering mechanism adopted by \cite{bergmann2019tracking} and \cite{zhou2020tracking}.
A new trajectory is initialized only if its bounding box at the current frame overlaps a public detection with IoU larger than 0.5.
We compare TransMOT-P with other trackers adopting the same filtering mechanism on the public detection track of MOT17 in Tab.~\ref{table:res17}. 
Compared with regular Transformer-based tracker~\cite{meinhardt2021trackformer}, TransMOT outperforms it in IDF1, MOTA, and IDS by a large margin. 
TransMOT also achieves the best IDF1 and MOTA scores among all published trackers, which demonstrates the robustness of TransMOT against detection quality variations.    

TransMOT-D adopts the DETR framework as detection and visual feature extraction sub-networks. 
TransMOT-D takes the detection outputs of pretrained TransTrack~\cite{sun2020transtrack} and their Transformer embedding as visual features.
For a fair comparison, the pretrained model of TransTrack is not fine-tuned in TransMOT-D.
We compare TransMOT-D and TransMOT with state-of-the-art trackers on MOT17 private detection track in  Tab.~\ref{table:res17}.
The performance of TransMOT-D is better than TransTrack~\cite{sun2020transtrack} by 10.0\% and 3.0\% in IDF1 and MOTA metrics respectively. 
This shows that our TransMOT framework can better model the spatial-temporal relationship of the tracklets and detections than standard Transformer. 
TransMOT-D achieves the lowest IDS among all the methods, because it generates fewer tracklets than other methods. 
Our TransMOT framework achieves the best IDF1 and MOTA metrics. 
It is also ranked as the top tracker in terms of IDF1 and FN on the private detector track of MOT17 challenge leaderboard when the paper is submitted.

\noindent\textbf{MOT20.}
MOT20 consists of eight sequences for pedestrian tracking. 
MOT20 video sequences are more challenging because they have much higher object density, \eg 170.9 \textit{vs} 31.8 in the test set. 
We report the experimental results of the proposed TransMOT and the comparison with the other methods in Tab.~\ref{table:res20p}. 
Our method establishes state-of-the-art in most metrics among all submissions.
In particular, TransMOT achieves 6.6\% higher IDF1 than CStrack, which ranked second in the leaderboard. 
In public detection setting, TransMOT-P also demonstrates its robustness to detection noise.
Compared with the private detector setting, the MOTA decreases 4.4\%, but IDF1 score only drops 0.9\%. 
The experiments on both public and private detection settings demonstrate that the capability of TransMOT of modeling a large number of tracklets and detections in crowd scenes as shown in Fig.~\ref{fig:vis}.


\subsection{Ablation}

We study the significance of different modules in the proposed method through ablation study as shown in Tab.~\ref{table:ab}. 
The ablations are conducted on the MOT17 \emph{training} set.
To avoid overfitting, the object detector used in the ablation study is trained on only the CrowdHuman dataset.

We first evaluate the settings that remove the Match with Motion module or the Long-Term Occlusion Handling module from the cascade association framework, noted as TransMOT$-$MwM and TransMOT$-$LTOH in Tab.~\ref{table:ab}. 
The MOTA score decreases significantly without MwM because the TransMOT cannot handle low score detection candidates well. 
It also leads to a slower speed, because TransMOT needs to handle more tracklets and detection candidates that were handled by MwM.
Dropping the LTOH module results in more fragmented tracklets because the tracklets that are occluded for more than $T$ frames are not linked in this case.
It dramatically increases the number of tracklets, reduces the IDF1 score, and increases the inference speed.

The major hyper-parameters of the tracker are also studied. 
We test TransMOT running at a reduced input image resolution at maximum width of 1280 (TransMOT@1280 in Tab.~\ref{table:ab}) (The default setting uses a maximum width of 1920).
We find that lower resolution input causes the tracker to miss small targets at the far end of the scene.
  The choice of the temporal history length $T$ is investigated by choosing $T=1$, $T=10$, and $T=20$ ($T$ is set to 5 in default setting). 
  Compared with $T=1$ where no temporal history is included, $T > 1$ can improve the association accuracy. 
  However, since increasing $T$ also adds more tracklets for association, it increases the complexity of the association task, and makes the learning harder under a limited number of training data for MOT. In addition, a lot of the tracklets that can be matched by increasing $T$ are already handled by the long-term occlusion handling module. Thus increasing $L$ beyond 5 has no performance gain and makes inference slower. 
  $T=5$ is used in all the other experiments.  

Finally, in addition to SiamFC and DETR features, we evaluate other shallow and deep visual features, including color histogram and ReID feature DGNet~\cite{zheng2019joint}.
Benefiting from the fully trainable Transformer, even using simple color histogram features, TransMOT can achieve similar performance with the one using deep ReID features and runs at a much faster inference speed. On the other hand, SiamcFC features perform better than both color histogram and ReID features, because it is trained on a large scale video dataset.

\begin{table}
	\footnotesize
	\begin{center}

		\begin{tabular}{c@{\hskip 3.1mm}|@{\hskip 3.5mm}c@{\hskip 3.5mm}c@{\hskip 3.5mm}c@{\hskip 3.5mm}}
			\hline\thickhline
			 Configuration  & IDF1 & MOTA & FPS   \\
			 \hline
			 TransMOT$-$MwM &  76.0 &  73.6 & 9.2\\
			 TransMOT$-$LTOH &  76.5 & 74.6  & 7.8\\
			 \hline
			 TransMOT@1280 & 75.3& 72.9 &  13.8\\
			 \hline
			 TransMOT@$T=1$ & 77.8& 74.7 & 9.8\\
			 TransMOT@$T=10$ & 78.0& 74.8 & 9.3\\
			 TransMOT@$T=20$ & 77.8& 74.7& 8.5\\
 			\hline
 			TransMOT+DGNet & 77.5& 74.7 & 6.0\\
            TransMOT+Histogram & 77.5 & 74.9 &  12.3 \\
			TransMOT+SiamcFC(\textbf{Ours}) & 78.1&74.8 &  9.6\\
			\hline\thickhline
		\end{tabular}

	\end{center}
	\caption{Ablations on the MOT17 benchmark training set. FPS indicates the inference speed of the full tracker that includes the detection and feature sub-net. }
    \label{table:ab}
\end{table}

\section{Conclusion}
We proposed a novel Spatial-Temporal Graph Transformer for multi-object tracking (TransMOT) with Transformers. 
By formulating the tracklets and candidate detections as a series of weighted graphs, the spatial and temporal relationships of the tracklets and candidates are explicitly modeled and leveraged. The proposed TransMOT not only achieves higher tracking accuracy, but also is more computationally efficient than the transitional Transformer-based methods.
Additionally, we developed the cascade association framework to further optimize the speed and accuracy of TransMOT by filtering low score candidates, recovering the tracklets that are occluded for a long time, and remove duplicate detections. Experiments on MOT15, MOT16, MOT17, and MOT20 challenge datasets show that the proposed approach achieves state-of-the-art performance on all the benchmark datasets.

{\small
\bibliographystyle{ieee_fullname}
\bibliography{egbib}
}

\end{document}